\immediate\write18{cp `kpsewhich apacite.bst` .}
\documentclass[twocolumn]{article}
\usepackage[english]{babel}
\usepackage{amsmath}
\usepackage{subcaption}
\usepackage{graphicx}
\usepackage{float}
\usepackage{caption}
\usepackage[colorlinks=true, allcolors=blue]{hyperref}
\usepackage{authblk}
\DeclareCaptionType{eq}[][List of equations]
\usepackage[nodayofweek,level]{datetime}
\newdate{date}{02}{05}{2022}
\date{\displaydate{date}}
\onecolumn
\title{CarbNN: A Novel Active Transfer Learning Neural Network
To Build De Novo Metal Organic Frameworks (MOFs) for Carbon Capture\\
  \large MATS055}
\author[1]{Neel Redkar\thanks{neel.redkar@gmail.com}}
\affil[1]{Independent Researcher — San Ramon CA, US}
\begin{document}
\maketitle
\begin{abstract}
Over the past decade, climate change has become an increasing problem with one of the major contributing factors being carbon dioxide (CO\textsubscript{2}) emissions—almost 51\% of total US carbon emissions are from factories. The effort to prevent CO from going into the environment is called carbon capture. Carbon capture decreases CO\textsubscript{2} released into the atmosphere and also yields steam that can be used to produce energy, decreasing net energy costs by 25-40\% \cite{rahimi_toward_2021}, although the isolated CO\textsubscript{2} needs to be sequestered deep underground through expensive means. Current materials used in CO\textsubscript{2} capture are lacking either in efficiency, sustainability, or cost \cite{unveren_solid_2017} \cite{rahimi_toward_2021}.

Electrocatalysis of CO\textsubscript{2} is a new approach where CO\textsubscript{2} can be reduced and the components used industrially as fuel, saving transportation costs, creating financial incentives. Metal Organic Frameworks (MOFs) are crystals made of organo-metals that adsorb, filter, and electrocatalyze CO\textsubscript{2}. The current available MOFs for capture \& electrocatalysis are expensive to manufacture and inefficient at capture \cite{rahimi_toward_2021}. Thus, the engineering goal for this project was to design a novel MOF that can adsorb CO\textsubscript{2} and use electrocatalysis to convert it to CO and O efficiently while maintaining a low manufacturing cost.

A novel active transfer learning neural network was developed, utilizing transfer learning due to limited available data on 15 MOFs \cite{shao_metal-organic_2020}. Using the Cambridge Structural Database with 10,000 MOFs, the model used incremental mutations to fit a trained fitness hyper-heuristic function \cite{groom_cambridge_2016}. Eventually, a Selenium MOF ($C_{18}MgO_{25}Se_{11}Sn_{20}Zn_{5}$) was converged on. Through analysis of predictions \& literature, the converged MOF was shown to be more effective \& more synthetically accessible than existing MOFs, showing the model had a understanding effective electrocatalytic structures in the material space. This novel network can be implemented for other gas separations and catalysis applications that have limited training accessible datasets.
\end{abstract}
\clearpage
\tableofcontents
\clearpage
\twocolumn
\begin{figure}
\centering
\includegraphics[width=0.4\textwidth]{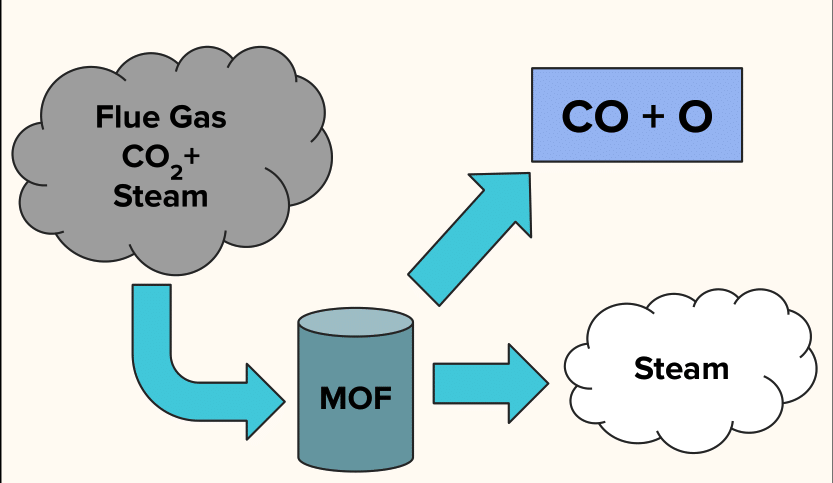}
\caption{\label{fig:g_abstract}A graphical representation of the MOF's (Metal Organic Framework) function.}
\end{figure}
\section{Introduction}
\subsection{Point-Source Carbon Capture as a Problem}
Atmospheric release of carbon dioxide from burning fossil fuels raises global temperatures and threatens to permanently damage the planet’s ecosystems.  One of the primary options to slow this is carbon capture, which prevents its emission at the source, such as a hydrocarbon-fueled power plan, and then reuses or store it. There are different modes of carbon capture, such as point-source and direct air. To keep ourselves carbon neutral, point-source carbon capture from factories in particular is key. Factories emit 51\% of total US carbon emissions, where point-source carbon capture increases costs 25-40\%, not being able to provide financial incentives \cite{unveren_solid_2017}. This is a major problem, because financial incentives are key to make large corporations make the shift over to have a neutral carbon footprint.

There are currently two viable adsorbants for point-source carbon capture, liquid amines (aqueous alkanolamine solutions) and solid adsorbants. Liquid amines are toxic to the environment and are unsustainable because they are volatile and must be constantly replenished \cite{unveren_solid_2017}. This makes them cost prohibitive.  The alternative, solid absorbents require less energy because they use pressure differentials for release of CO\textsubscript{2} \cite{unveren_solid_2017}. Development is ongoing into creating solid adsorbents that are able to adsorb CO\textsubscript{2} efficiently.

\subsection{Metal-Organic Frameworks (MOFs)}
\subsubsection{Metal-Organic Frameworks}
Metal-Organic Frameworks are organo-metals joined by organic ligands that can have an assortment of properties. The complex properties that can arise from this 3-dimensional yet simple structure makes it a great candidate for a variety of uses. Their extremely high surface area (porosity) also make them promising choices for solid adsorbants of CO\textsubscript{2} \cite{unveren_solid_2017}. Research is being conducted in the adsorbance of CO\textsubscript{2} for the capture of carbon. 

\subsubsection{Electrocatalysis Benefits}
Electrocatalysis is another use for MOFs \cite{shao_metal-organic_2020}. The ligand sites have the ability to convert CO\textsubscript{2} into carbon monoxide and oxygen. Outputs of the reaction can individually be used for fuel and industrial oxidization reactions \cite{keim_carbon_1989} \cite{zheng_recent_2018}. This provides a further financial incentive by producing a usable byproduct from the capture. Current carbon capture calls for sequestration of CO\textsubscript{2}, which requires extra costs as well as large pipelines to be built underground to place it pressurized under deep "caprock" or a layer that prevents the air from leaking to the surface. By catalyzing it into usable byproducts, savings can be made in elimination of the sequestration as well as selling/repurposing concentrated carbon monoxide.
\begin{figure*}[t!]
\centering
\includegraphics[width=0.76\textwidth]{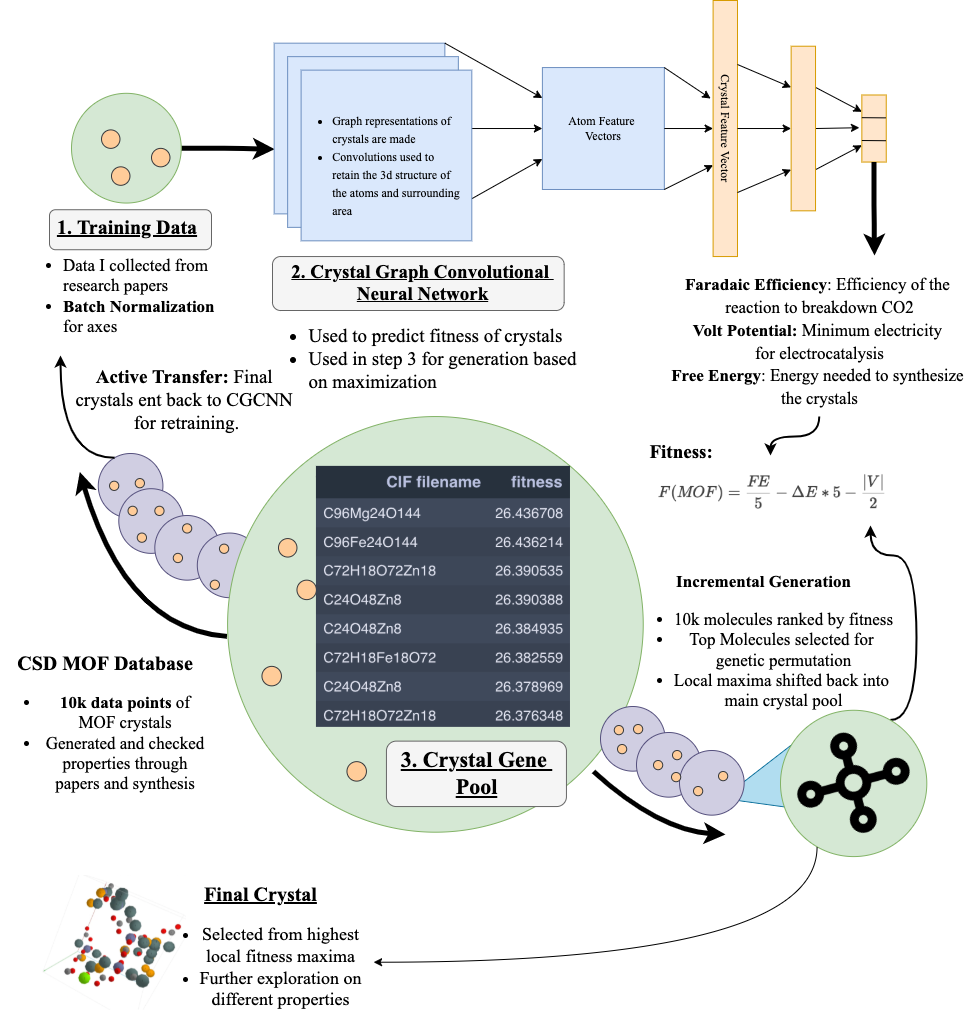}
\caption{\label{fig:model}Schematic of the process used to generate novel MOFs. Starting with the training data (1) it runs through crystal graph convolutional neural networks (2) to correlate values for predictions. These are aggregated in a hyper-heuristic fitness function, then evolution used to maximize the fitness function (3). Maxima are brought back into training data (1) for active transfer learning.}
\end{figure*}
\begin{table*}[t!]
\caption {Fitness Function Parameters} \label{tab:fitness_table} 
\begin{center}
\begin{tabular}{ |c|p{8cm}| } 
 \hline\hline
 Faradaic Efficiency ($FE$) & Efficiency of the reaction to breakdown CO\textsubscript{2} \cite{shao_metal-organic_2020}. \\ 
 \hline
 Voltage Potential ($V$) & Minimum electricity needed for electrocatalysis \cite{shao_metal-organic_2020}. \\ 
 \hline
 Free Energy ($\Delta E$) & Energy needed to synthesize crystals \cite{shao_metal-organic_2020}. Commonly correlated to the synthetic accessibility and cost of the crystal \cite{anderson_large-scale_2020}. \\ 
 \hline
\end{tabular}
\end{center}
\end{table*}
\subsection{Machine Learning Architectures}
Since the MOF space is diverse with many possible properties, which makes exploration key to finding substances that match the attributes that one is looking for. Current methods are highly reliant on experimentation where they need to guess and test possible MOFs. This leads to staying close to the known, as well as often missing radically novel MOFs that could function better. Machine learning is a solution which is efficient at taking large feature spaces and searching for maxima. This has been a common method for finding novel MOFs for different processes \cite{chong_applications_2020}. However, most of these methods use high throughput screening. This approach uses large amounts of data, while most reactions only have very low amounts of experimental data. Common methods utilized are Monte Carlo trees and generative adversarial neural networks, both of which use large amounts of data—10K+ \cite{chong_applications_2020} \cite{zhang_machine_2021}, as well as fail to account for spatial attributes. Monte Carlo trees are usually promising in such tasks, but the making the structure of MOFs linear to fit a tree loses essential data that can hurt end products \cite{zhang_machine_2021}. Architectures such as the ones outlined have been utilized, but the largest flaw is that neural networks either don't explore the space well, or do not function well with limited data \cite{zhang_machine_2021}. This is especially detrimental because a majority of niche important tasks have only a handful of tested MOFs published.

\section{Engineering Goal}

The engineering goal for this paper was to use machine active-transfer learning to create and optimize a novel MOF to capture carbon \& have ligand sites that induce electrocatalysis. To test effectiveness, it should demonstrate that the novel MOF has a higher electrocatalytic ability than published options and is synthetically accessible/reduced cost. Lastly the framework shown should allow for interoperability and easy addition of data and new properties for continued optimization.
\subsection{Active Transfer Learning}
Active transfer learning was used because of its ability to work well with limited amounts of data. By exploring the unknown space slowly, it can open up unique possibilities that researchers weren't able to search before. Since it also explores slowly, the predictions should be more accurate than other methods, slowly correcting itself \cite{kim_deep_2021}. The reason for this is because it gets to correct the maxima that it thought were accurate, along with fixing clear errors in the algorithm direction.
The way active transfer learning does this is by taking the maxima of evolution and putting it back into the initial training data for a fitness function. This way it expands the known space iteratively \cite{kim_deep_2021}. Different data augmentation techniques can be used for the evolutionary algorithm, but the insertion of maxima back into the training data remains the same. This can also be seen in the gene pool (3) in Figure \ref{fig:model}.2. This type of algoritm also allows for wet lab tests to be done for key points in the dataset, which makes new additions to training data post-synthesis more valuable.
\section{The Novel Algorithm}
\subsection{Data Gathering}
Data was gathered for the electrochemical reaction below:
$$2CO_{2} \longrightarrow 2CO + O_{2}$$
Data was gathered through searching of various databases for MOFs that had the specific electrochemical properties. The main properties were decided for ease of the electrocatalytic reaction, as well as probability for efficient synthesis. The variables are referenced in the Table \ref{tab:fitness_table}, and data was gathered through Shao (2020)'s summary of the electrochemical space for the reduction of CO\textsubscript{2} \cite{shao_metal-organic_2020}. Free energy was adapted from Anderson (2020) finding significant correlations with lowering the free energy and the synthetic accessibility of the MOF \cite{anderson_large-scale_2020}. All data was gathered into CIF (Crystallography Information Framework) files that can be found at the repository. CIFs accounts for spatial dimensions \& angles that get used in the neural network, as opposed to SMILES (Simplified Molecular-Input Line-Entry System) or other encoding methods.
\subsection{Fitness and Regression}
\begin{figure*}[t!]
    \centering
    \includegraphics[width=0.6\textwidth]{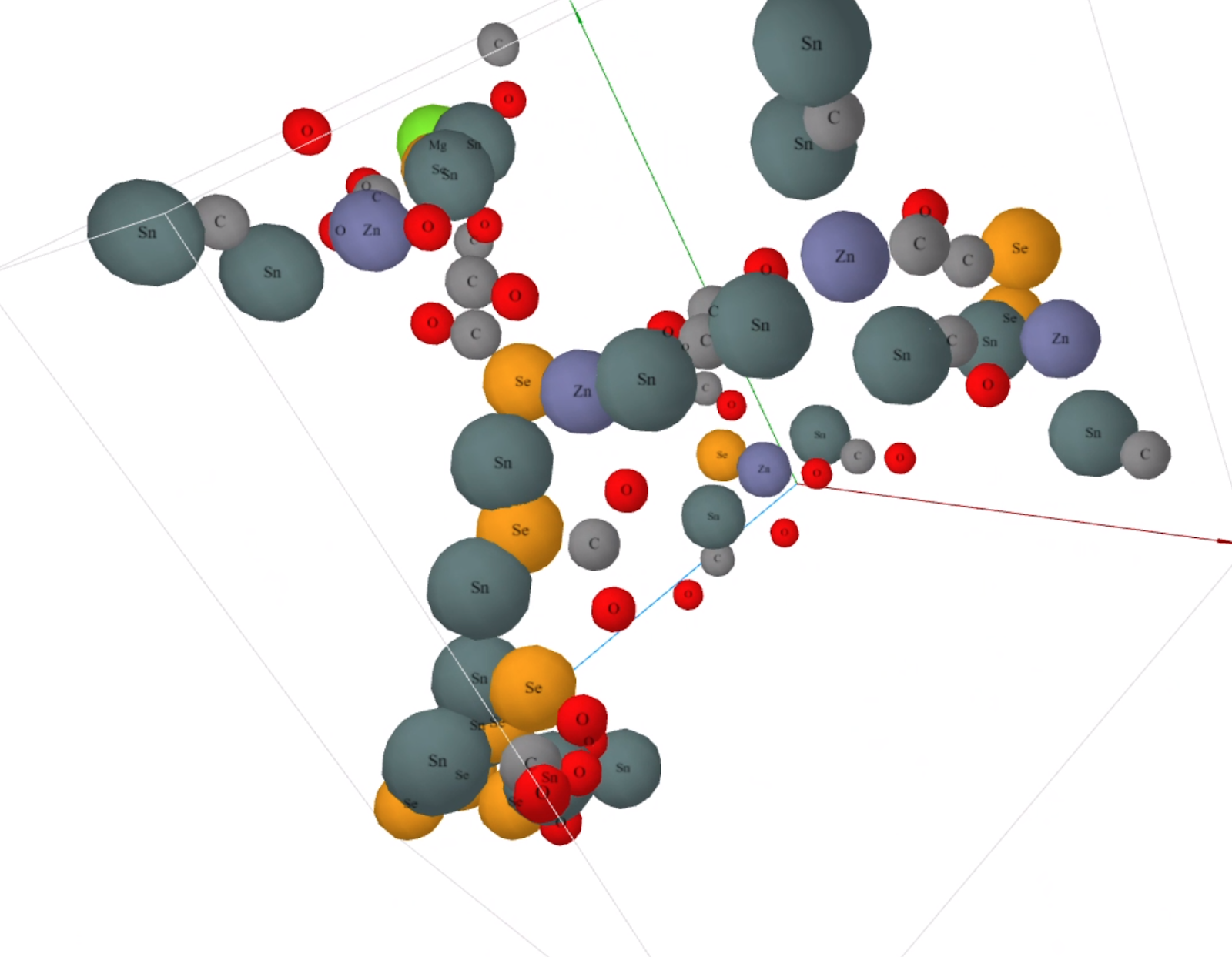}
    \caption{Converged MOF $C_{18}MgO_{25}Se_{11}Sn_{20}Zn_{5}$}
    \label{fig:converged_mof}
\end{figure*}
The models used for the fitness function were Crystal Graph Convolutional Neural Networks (CGCNNs) trained on 15 gathered molecules. The network was adapted from Xie (2017) which has been used in the area of crystals \cite{xie_crystal_2018}. In this model, crystals are turned into undirected graphs and then convolved upon to get feature vectors. This is more efficient than linear vectors as it preserves the spatial directions as well as intermolecular interactions. In the article they were found to have a significantly close accuracy to other models in the area that were trained with pretested values \cite{xie_crystal_2018}. The initial undirected graph to feature vectors can also use transfer learning as inter-molecular reactions would ideally stay the same between crystals. As MOFs are crystals in structure, this model seemed to be the best for MOFs without any prior data, as well as novel generated MOFs.

New models were created via training CGCNNs on one property each, one for faradaic efficiency and one for voltage potential seen in Figure \ref{fig:model}. Free energy was taken as a pretrained model using data from formation energy reactions in general crystals \cite{jain_commentary:_2013} \cite{xie_crystal_2018}. These three models were brought together to create a hyper-heuristic fitness function that was modified to normalize all three values. Most values were fairly arbitrary to normalize them and produce viable crystals, so this would be an ideal place for future research.
\begin{eq}
\caption{Variable definitions in Table \ref{tab:fitness_table}}
$$Fitness(MOF) = \frac{FE}{5} - \Delta E * 5 - \frac{|V|}{2} $$
\end{eq}
\subsection{Iterative Evolution/Active Transfer Learning}
As a base for the evolutionary algorithm, 10K MOFs were taken from the Cambridge Structural Database \cite{groom_cambridge_2016}. These were pregenerated with reasonable integrity, and then ranked via the fitness function described previously. The top 100 data points were augmented using different data augmentation techniques. Certain probabilities were decided between full structure mutations (to test new scaffolding), new atom additions/replacements (to test new materials), as well as slab crossovers (to simulate crossing of genes in nature) \cite{hjorth_larsen_atomic_2017}. These were all utilized to simulate biological evolution, as well as extra variation to test new atoms for the framework. 

Many problems were run into during the evolution simulation due to the complex structure of the crystals and need to make crossover permutations function between crystals with fundamentally different structures. Different methods were used to edit structures to fit different axis alignment requirements. 

Active transfer learning was then used when bringing the peaks of the evolution back into the initial CGCNN fitness function training dataset. This was done with the predicted values to iteratively increase the known space, as well as adjusted to approximate values. Iterative exploration with data augmentation/mutation allows for very slow expansion of the learned space, which leads to less errors, as opposed to predictions far away from the learned. The increase in effectiveness can also be attributed to the fixing of glaring errors during active transfer learning (no organic linkers, unfinished structures etc.), which led to greater accuracy. This can be seen in figure \ref{fig:model} (3).
\begin{figure*}[t!]
    \centering
    \begin{subfigure}[b]{0.475\textwidth}   
        \centering 
        \includegraphics[width=\textwidth]{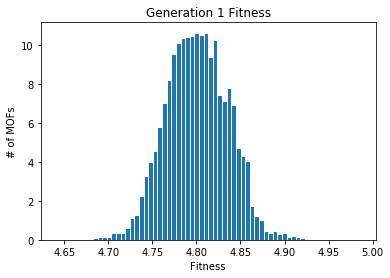}
        \caption[Gen 1 Fitness]%
        {{\small Generation 1 Fitness Distribution}}    
        \label{fig:gen_1}
    \end{subfigure}
    \hfill
    \begin{subfigure}[b]{0.475\textwidth}   
        \centering 
        \includegraphics[width=\textwidth]{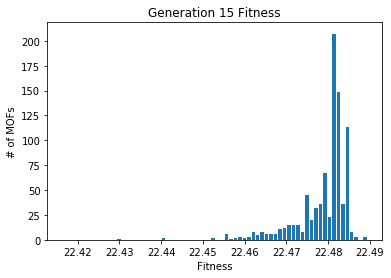}
        \caption[Gen 15 Fitness]%
        {{\small Generation 15 Fitness Distribution}}    
        \label{fig:gen_15}
    \end{subfigure}
    \vskip\baselineskip
    \begin{subfigure}[b]{0.475\textwidth}
        \centering
        \includegraphics[width=\textwidth]{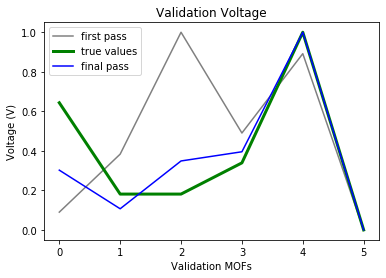}
        \caption[Voltage Validation]%
        {{\small Validation Voltage MOF Graph (x axis discrete)}}    
        \label{fig:voltage_validation}
    \end{subfigure}
    \hfill
    \begin{subfigure}[b]{0.475\textwidth}  
        \centering 
        \includegraphics[width=\textwidth]{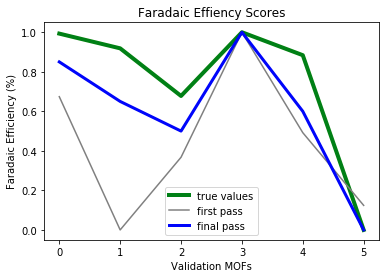}
        \caption[Faradaic Efficiency Validation]%
        {{\small Validation Faradaic Efficiency MOF Graph (x axis discrete)}}    
        \label{fig:fara_eff_validation}
    \end{subfigure}
    \caption[]%
    {\small Training graphs show that the model had succeeded in learning} 
    \label{fig:graphs}
\end{figure*}
\subsection{Training}
The model did succeed in training and did not over-fit values due to the active transfer learning. The fitness shifted from a mean of around 4.8 in the generation 1 graph \ref{fig:gen_1} to 22.48 in the 15th generation \ref{fig:gen_15}. The peaks of that generation were then loaded back into the training dataset.

Through active transfer learning, the model was able to even out major differences in the model. This is shown through the validation MOFs which were not shown to the model. Although for validation, the voltage potential values were off by quite a bit on the first pass of training, after more data was added to the training dataset, it started to converge \ref{fig:voltage_validation}. This was also shown for the Faradaic efficiency evening out substantial differences in the percentage \ref{fig:fara_eff_validation}. For reference, all values were normalized between 0-1.
\begin{figure*}[t!]
    \centering
    \begin{subfigure}[b]{0.475\textwidth}   
        \centering 
        \includegraphics[width=0.9\textwidth]{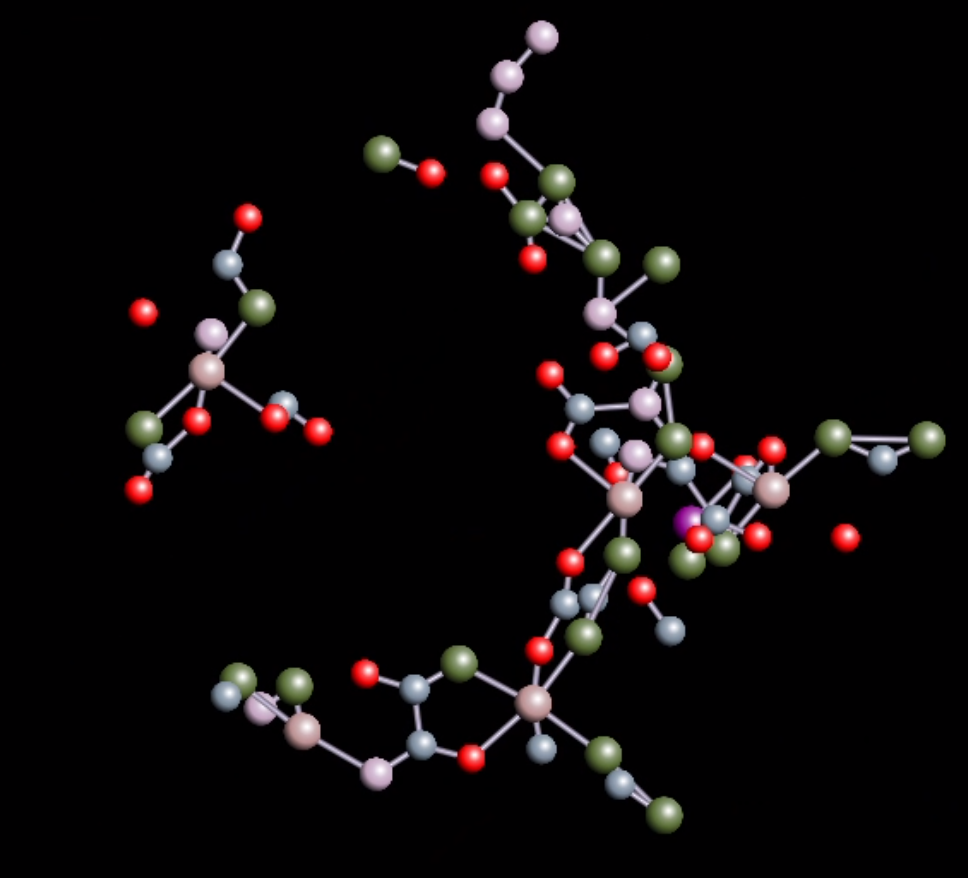}
        \caption[black background converged mof]%
        {{\small Converged MOF Alternate Diagram}}    
        \label{fig:alt_mof_conv}
    \end{subfigure}
    \vskip\baselineskip
    \begin{subfigure}[b]{0.475\textwidth}
        \centering
        \includegraphics[width=0.2\textwidth]{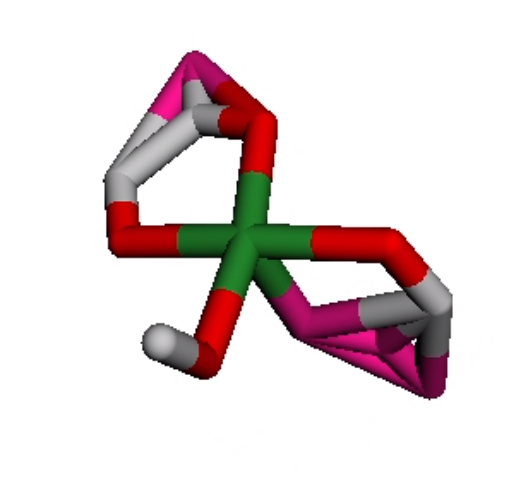}
        \caption[Repeated Metal]%
        {{\small Converged MOF Repeated Metals}}    
        \label{fig:converged_metals}
    \end{subfigure}
    \hfill
    \begin{subfigure}[b]{0.475\textwidth}  
        \centering 
        \includegraphics[width=0.2\textwidth]{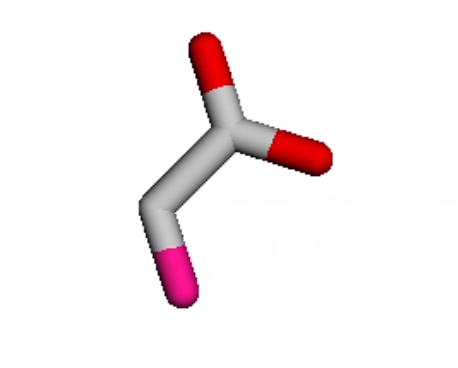}
        \caption[ligand]%
        {{\small Converged MOF Ligands}}    
        \label{fig:converged_ligands}
    \end{subfigure}
    \caption[]%
    {\small The converged MOF parts \& diagrams $C_{18}MgO_{25}Se_{11}Sn_{20}Zn_{5}$} 
    \label{fig:mols}
\end{figure*}
\section{Converged Results}
The converged MOF structure can be seen in Figure \ref{fig:converged_mof}. The base structure of the molecule is $C_{18}MgO_{25}Se_{11}Sn_{20}Zn_{5}$. Overall fitness of the MOF was ~32 with a faradaic efficieny of 99.99\%, voltage potential of 11.26V, and free energy of -3.39 eV/atom. The higher FE, lower voltage potential, and lower free energy shows that the evolution algorithm worked, even though typically these algorithms tend to overfit. None of the prior MOFs seemed extremely similar to the generated MOF, which indicates that it used its learned intermolecular reactions from the small dataset onto the CSD pool.
\ref{fig:converged_metals} individually.
\begin{figure}[H]
\centering
\includegraphics[width=0.49\textwidth]{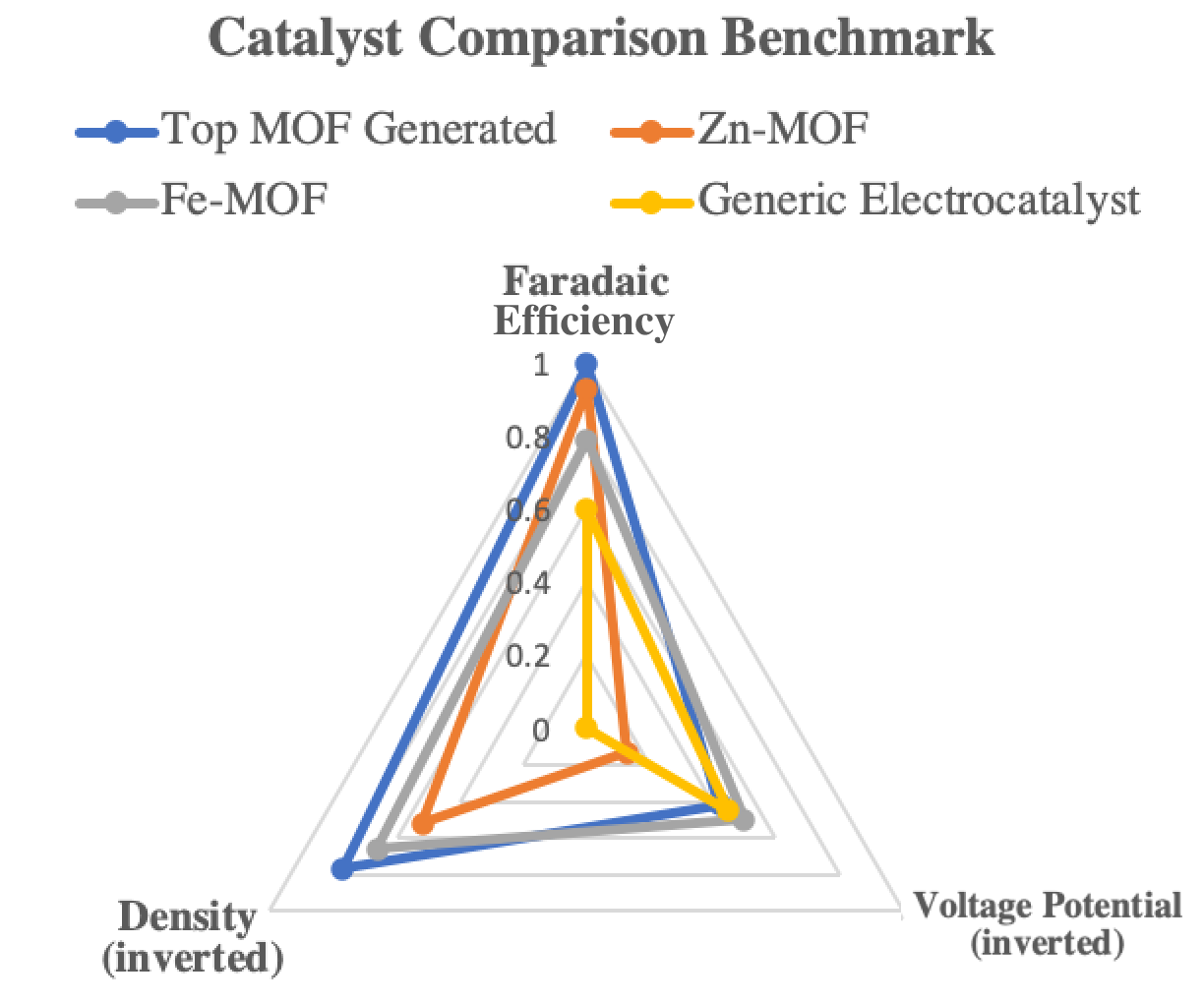}
\caption{\label{fig:radar}Radar chart comparison of prior top catalysts (Zn-MOF \cite{van_phuc_highly_2021}, Fe-MOF \cite{huan_electrochemical_2017}, and non-MOF catalysts \cite{shen_electrocatalytic_2015}) to the generated one.}
\end{figure}

Figure \ref{fig:radar} shows a radar graph which is useful to determine for comparison of substances with many features \cite{malek_data-driven_2021}. The converged MOF is more efficient and less dense than other alternative MOFs, meaning that it could convert more electricity per amount of CO\textsubscript{2}, while having larger amounts of passthrough than other MOFs. The closest in terms of area are Fe-MOFs \cite{huan_electrochemical_2017}, but there is a 21\% decrease in faradaic efficiency compared to the generated MOF, with a more conservative estimate being ~15\%. These are substantially greater than prior MOFs, especially with the low voltage potential implying it low power.

In the paper, we have not been able to synthesize this MOF due to not having access to financial resources. As an attentive evaluation, an analysis was performed of the makeup of the MOF including features in the ligands \ref{fig:converged_ligands} and the metals. Selenium based MOFs are commonly used in batteries and is correlated with high conductivity and porosity \cite{li_preparation_2020}. Conductivity is key for an electrocatalytic reaction to be efficient, along with porosity needed for carbon capture and storage. Magnesium ligands are seen throughout the MOF and have been seen to be extremely successful in oxygen reduction reactions \cite{liu_turning_2020}. The bonds are similar for CO\textsubscript{2} which indicates that it might have found novel structures that could be used for reduction. Lastly, the zinc build of the MOF is common for carbon capture, and shows that the MOF has a firm basis in the current literature. All of these put together, show that through active transfer learning, the model picked out these key properties that would be essential for carbon capture \& electrocatalysis. With the model selecting these specific attributed from nothing, it has shown that active transfer learning did work with extremely low data. Magnesium and selenium structures were also not seen in past data, which indicates the maximization worked. 

The final MOF was also tested using the MIT MOFSimplify benchmark that yielded a structural integrity of 88\% \cite{nandy_mofsimplify_2022}. This is relatively high, and indicative that it would be able to be synthesized as well as the structure being able to be stable after removal of the solvent during synthesis. The thermal stability was predicted to breakdown at 210\textdegree  C, which for the task inside of factories is reasonable, especially for direct air capture.

\section{Conclusion}
\subsection{Discussion}
In conclusion, the engineering goal was achieved and the model did achieve high accuracy as shown in graphs \ref{fig:voltage_validation} and \ref{fig:fara_eff_validation}. The properties of the MOF were shown to match those expected, as well as new promising possibilities for electrocatalysis.

The MOF converged upon had a higher FE than prior MOFs with an approximate 7-19\% increase in efficiency as well as being more synthetically accessible than prior MOFs. With the lowest free energy, this would be a good predictor of ease to synthesize, making it a possibly less expensive alternative in manufacturing costs (though processes would be unknown due to manufacturing methods being complex \& to each product). Other parameters that might be needed to added to the model in the future (heat stability etc.) can be implemented into the fitness function with relative ease. It would be difficult to calculate the exact embodied CO\textsubscript{2} savings due to how far it is from manufacturing capacity, but relative to current options it should be approximate to the faradaic efficiency difference. Current capture is inefficient with ~50\% efficiency, and MOF use would decrease vaporized waste, energy costs (pressure swing adsorption is efficient), and provide usable fuel from the carbon to decrease costs further in addition to the predicted efficiency and voltage potential savings \cite{unveren_solid_2017}\cite{university_2019}.

The model also worked exceedingly well with low initial data. Being able to identify areas important for carbon capture with low amounts of data is impressive. The inclusion of free energy calculations in the model was unique for generation models in the MOF field, which has also proven to work effectively to generate novel molecules. The model is also open source and built to be interoperable with many fitness functions.

This active transfer learning model would benefit to a greater extent in a lab environment where testing of the MOF could be done to gain highly accurate results to correct the network. This would mean a failure to synthesize wouldn't be detrimental to the network, but help guide the network toward a global maxima. To speed up feedback loops for experimentation the fitness function could be changed to place more emphasis on ease of synthesis in a lab setting.
\subsection{Industrial Methods}
The specific industrial architecture for the MOF is not in the scope of this paper, though a common method that would be used is Pressure Swing Adsorption (PSA) \cite{unveren_solid_2017}. This method would utilize multiple MOFs in a rotating disc to capture carbon dioxide out of outgoing flue gas. Electricity would be running at ideally ~11V through the MOF to catalyze the CO\textsubscript{2} into CO and O\textsubscript{2}. These would then be brought into a low-pressure chamber where the gas would leak out of the MOF.

The low voltage potential would allow it to be run by renewable energy \cite{zheng_recent_2018} places for direct air capture. By using renewable energy to convert CO\textsubscript{2} into fuel, that would then be converted back into CO\textsubscript{2}. This would create a closed circle loop carbon that could be utilized for energy, that is powered by excess renewable energy. This would be direct air carbon capture, but if the MOF predicted is successful, it could be utilized in such tasks. Hopefully the MOF could provide a financial transition between point source carbon capture, into direct air carbon capture, which would then be utilized for this sustainable model of energy storage in CO fuel (described in Zheng) \cite{zheng_recent_2018} \cite{ruiz-lopez_electrocatalytic_2022}. 

Future work would need to be done on separation and purification of CO and O\textsubscript{2} for industrial use. Once done, this would enable the use of CO in oxidization reductions and fuel \cite{keim_carbon_1989} \cite{zheng_recent_2018}. The O\textsubscript{2} could be released for an environmentally positive effect, fuel, or medical use.
\subsection{Future Work}
\subsubsection{Converged MOF Use Cases}
If successful in electrocatalysis after synthesis of the MOF, this approach would provide large financial incentives for factories to switch over to become carbon neutral. The MOF would be able to cut sequestration out of the carbon capture process, getting rid of active pipelines and pumping stations. The fuel could also turn CO\textsubscript{2} into a net positive resource, providing financial incentives to turn green decreasing cost for consumers. The O\textsubscript{2} could be released into the environment as a net positive or also be put back into industrial use. This realistic view into company financials and carbon reusability is essential to become carbon neutral without destroying factories.
\subsubsection{The Novel Algorithm}
The algorithm has also been proven to work exceedingly well with low data. Cross application into different catagories would be significant, due to the majority of MOF uses having only a handful of data points. Possibilities include photocatalysis, water treatment, and minimal data gas separation tasks \cite{chen_metalorganic_2017}. Researchers have reached out and future work might be done in their mentorship, as well as possible further synthesis.

Key areas for model improvement in the fitness function is inclusion of elements like specific heat along with other factors that contribute to more real world desirable attributes. Gathering negative controls/failed experiments is likely to also prove beneficial due to giving networks nuance into close structures that do not work \cite{moosavi_capturing_2019}. This would include contacting labs that synthesized successfully for their failed experiments to gather.
\subsection{Graphs/Figures}
All graphs and figures were created and generated by the researcher.
\subsection{Data Availability Statement}
The data that support the findings of this study are openly available in the GitHub at \href{https://github.com/neelr/carbnn}{https://github.com/neelr/carbnn}, reference number \cite{github}.
\nocite{*}
\bibliographystyle{plain}
\bibliography{main}

\end{document}